\documentclass[letterpaper, 10 pt, conference]{ieeeconf}  
\usepackage{cite}
\usepackage{amsmath,amssymb,amsfonts}
\usepackage{graphicx}
\usepackage{hyperref} 
\usepackage{textcomp}
\usepackage{tabularx}
\usepackage{amsmath}
\usepackage{balance}
\usepackage{multirow}
\usepackage{dblfloatfix}
\usepackage{siunitx}
\usepackage{booktabs}
\usepackage{longtable}
\usepackage{cite}
\usepackage{color,colortbl}
\usepackage{algpseudocode}
\usepackage{algorithm}
\definecolor{Gray2}{gray}{0.9}
\definecolor{Gray}{gray}{0.7}

\usepackage{xcolor}
\usepackage[font=small]{caption}
\usepackage{titlesec}
\usepackage{subcaption}
\titlespacing*{\section}
{0pt}{1.2ex plus 0.5ex minus 0.2ex}{0.8ex}

\titlespacing*{\subsection}
{0pt}{1.0ex plus 0.5ex minus 0.2ex}{0.6ex}

\titlespacing*{\subsubsection}
{0pt}{0.8ex plus 0.3ex minus 0.1ex}{0.4ex}

\IEEEoverridecommandlockouts                              

\usepackage{graphicx} 

\title{\LARGE \bf
AgenticDiffusion: Multi-View Reasoning with View-Conditioned Diffusion Planning for Vision-Based UAV Navigation
}
\author{Faryal Batool, Muhammad Ahsan Mustafa, Fawad Mehboob, Valerii Serpiva, and Dzmitry Tsetserukou}
\begin{document}

\maketitle
\thispagestyle{empty}
\pagestyle{empty}

\begin{abstract}

Vision-based UAV navigation becomes challenging when navigation targets are distributed across complementary camera views and cannot be reliably observed from a single viewpoint. We propose \textbf{AgenticDiffusion}, an agentic multi-view UAV navigation framework that semantically coordinates first-person-view (FPV) and top-view observations for mission-level navigation. Given a natural-language instruction, AgenticDiffusion identifies the requested targets, selects the most appropriate camera view for each navigation task, determines the corresponding navigation goal, and invokes the appropriate view-conditioned diffusion planner for trajectory generation. The resulting trajectories are executed using Nonlinear Model Predictive Control (NMPC).

AgenticDiffusion was evaluated in four real-world indoor scenarios, achieving an overall mission success rate of \textbf{80\% across 40 physical-flight trials}. In mixed-visibility scenarios, where the requested targets were distributed across FPV and top-view observations, coordinated multi-view navigation reduced average mission time by \textbf{50.8\%} relative to FPV-only navigation and by \textbf{26.8\%} relative to Top-only navigation. The semantic view-selection mechanism was also robust to lexical variation in target descriptions, achieving \textbf{100\% accuracy across 66 test cases}, compared with \textbf{63.64\%} for a confidence-based view-selection baseline. In a substantially larger Gazebo environment, AgenticDiffusion achieved a \textbf{90\% mission success rate} and completed the multi-stage mission, whereas the FPV-only and Top-only variants were unable to complete all requested navigation tasks.

\end{abstract}


\section{INTRODUCTION}

Autonomous UAV navigation in indoor environments requires a robot to interpret semantic task objectives while accounting for obstacle configurations, target visibility, and scene accessibility. Many vision-based navigation systems rely primarily on observations from a single viewpoint, which can limit their ability to handle occlusions or reason about scene structure beyond the currently visible region. Consequently, targets that are poorly represented from one viewpoint may require additional observation acquisition and repeated semantic inference before a feasible navigation goal can be determined. Recent advances in learning-based navigation, including reinforcement learning~\cite{navrl}, diffusion-based planning~\cite{navdp,nomad}, and vision-language navigation~\cite{navid}, have demonstrated strong potential for autonomous robot navigation. However, these approaches generally do not explicitly coordinate complementary camera perspectives to determine which view is most appropriate for a given semantic navigation task.

\begin{figure}[t]
\centering
\includegraphics[width=1\linewidth]{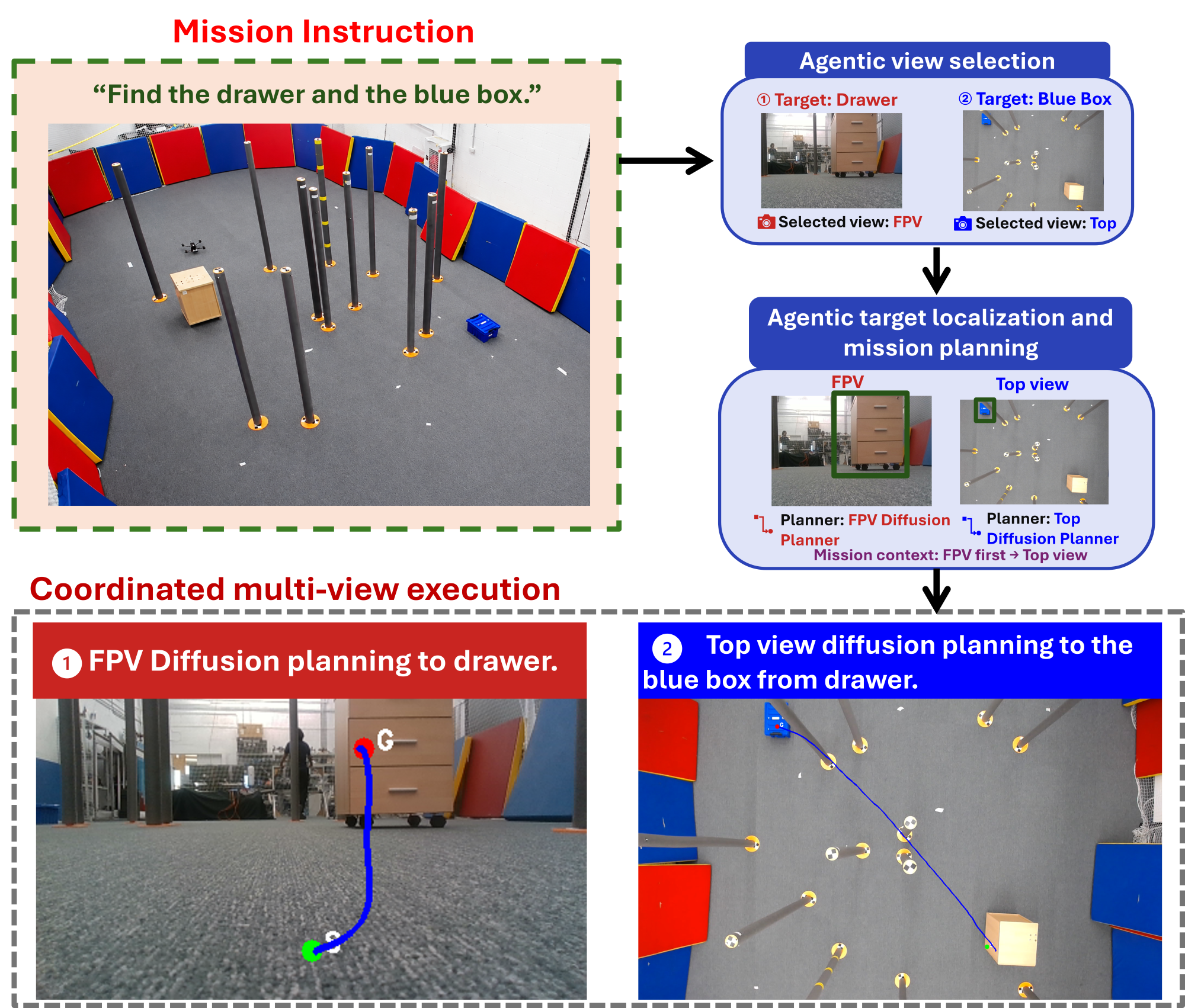}
\caption{Multi-view semantic reasoning and mission-level navigation in AgenticDiffusion. The framework associates requested targets with suitable camera views, determines navigation goals, and coordinates the corresponding view-conditioned planners for execution.}
\label{fig:main_figure}
\end{figure}

To address this limitation, we propose \textbf{AgenticDiffusion}, an agentic multi-view framework for semantic UAV navigation. Given a natural-language instruction and available first-person-view (FPV) and top-view observations, AgenticDiffusion identifies the requested targets, selects the most appropriate camera view for each navigation task, and determines the corresponding navigation goals. The framework then reasons over the resulting target--view associations together with the UAV state and camera configuration to determine the task execution order and select the appropriate view-conditioned diffusion planner for each task. The generated trajectories are executed sequentially using nonlinear model predictive control (NMPC).

By coordinating complementary camera viewpoints at the mission level, AgenticDiffusion can avoid unnecessary observation acquisition and repeated semantic reasoning when different targets are better represented from different camera perspectives, as illustrated in Fig.~\ref{fig:main_figure}. This enables a single natural-language mission to be decomposed into view-specific navigation tasks while retaining a unified execution strategy.

The main contributions of this paper are:

\begin{itemize}

\item We propose \textbf{AgenticDiffusion}, an agentic multi-view UAV navigation framework that coordinates FPV and top-view observations to associate each requested target with an appropriate camera view and corresponding view-conditioned diffusion planner.

\item We introduce a mission-level reasoning strategy that interprets natural-language instructions, determines navigation goals, and uses the resulting target--view associations to sequence multiple navigation tasks for execution.

\item We evaluate AgenticDiffusion in four real-world indoor UAV scenarios, examining multi-view navigation efficiency and robustness of semantic view selection, and additionally assess the framework in the substantially larger Gazebo Small City environment.

\end{itemize}

\section{Related Works}

Autonomous UAV navigation in indoor environments remains challenging due to occlusions, restricted fields of view, dynamic obstacles, and ambiguity in semantic target localization. Recent advances in learning-based planning, embodied reasoning, and vision-language navigation have improved autonomous decision-making and trajectory generation. However, many existing approaches rely on a single or fixed observation modality and do not explicitly coordinate complementary viewpoints for mission-level navigation.

\subsection{Learning-Based and Diffusion Navigation Frameworks}

Learning-based navigation methods have demonstrated strong performance in obstacle-aware navigation and long-horizon trajectory generation~\cite{navrl,3drldwa,macroactions}. More recently, diffusion-based approaches such as NoMaD~\cite{nomad}, NavDP~\cite{navdp}, and related trajectory-generation frameworks~\cite{dipper,dtg,NavDiffuser} have shown promising results for visual navigation and mapless planning. These methods learn to generate navigation trajectories directly from sensory observations and can improve robustness in cluttered environments. However, they generally assume a predefined observation modality and do not explicitly reason about which viewpoint should be used for a given semantic navigation task. They also typically do not perform mission-level selection among view-conditioned planners for aerial navigation.

\subsection{Multi-View and Spatial Navigation Frameworks}

Multi-view perception and navigation have been explored as means of improving environmental awareness under viewpoint-limited observations. Prior work includes multi-view reinforcement learning~\cite{10.5555/3507788.3507815}, multi-camera UAV perception and SLAM systems~\cite{YANG2017116,9341304}, and viewpoint-aware navigation frameworks~\cite{10.5555/3600270.3600510,xu2025mmnavmultiviewvlamodel}. These approaches demonstrate the value of combining information from multiple visual perspectives, but their primary focus is typically perception, representation learning, or navigation policy design rather than semantic task decomposition across complementary views.

In parallel, language-grounded spatial representations such as VLMaps~\cite{huang2023visuallanguagemapsrobot}, SpatialVLM~\cite{10658310}, and MapNav~\cite{zhang2026mapnavnovelmemoryrepresentation} improve spatial reasoning and scene understanding for embodied agents. Memory-guided exploration and synthetic vision-language navigation training methods~\cite{lee2025imagineverifyexecutememoryguided,11063767} further support semantic exploration in partially observed environments. While these methods improve spatial and semantic reasoning, they do not directly address the problem of associating different mission targets with complementary camera viewpoints and routing those targets to corresponding vision-conditioned trajectory planners in a UAV navigation pipeline.

\subsection{Agentic Vision-Language Navigation Frameworks}

Recent embodied AI systems increasingly use vision-language models (VLMs), large language models (LLMs), and agentic reasoning for instruction-guided navigation and task planning. Vision-language navigation systems such as NaVid~\cite{navid}, Fast-SmartWay~\cite{fastsmartway}, JanusVLN~\cite{janusvln}, and related embodied navigation agents~\cite{agentvln,openuav} have demonstrated the ability to interpret natural-language instructions and reason about semantic navigation objectives. Agentic systems such as Voyager~\cite{wang2023voyageropenendedembodiedagent}, SMART-LLM~\cite{10802322}, hierarchical UAV agents~\cite{herron2025hierarchicalagenticframeworkautonomous}, and FlowAgent~\cite{flowagent} further illustrate the use of hierarchical reasoning and generative decision-making for embodied tasks.

AgenticDiffusion differs from these approaches by combining mission-level semantic reasoning across complementary FPV and top-view observations with view-conditioned diffusion planning and NMPC-based UAV execution. Rather than relying on a single visual stream, the framework associates each requested target with an appropriate camera view, determines the corresponding navigation goal, and assigns the task to the relevant diffusion planner before trajectory execution.

\begin{figure*}[t]
\centering
\includegraphics[width=1\linewidth]{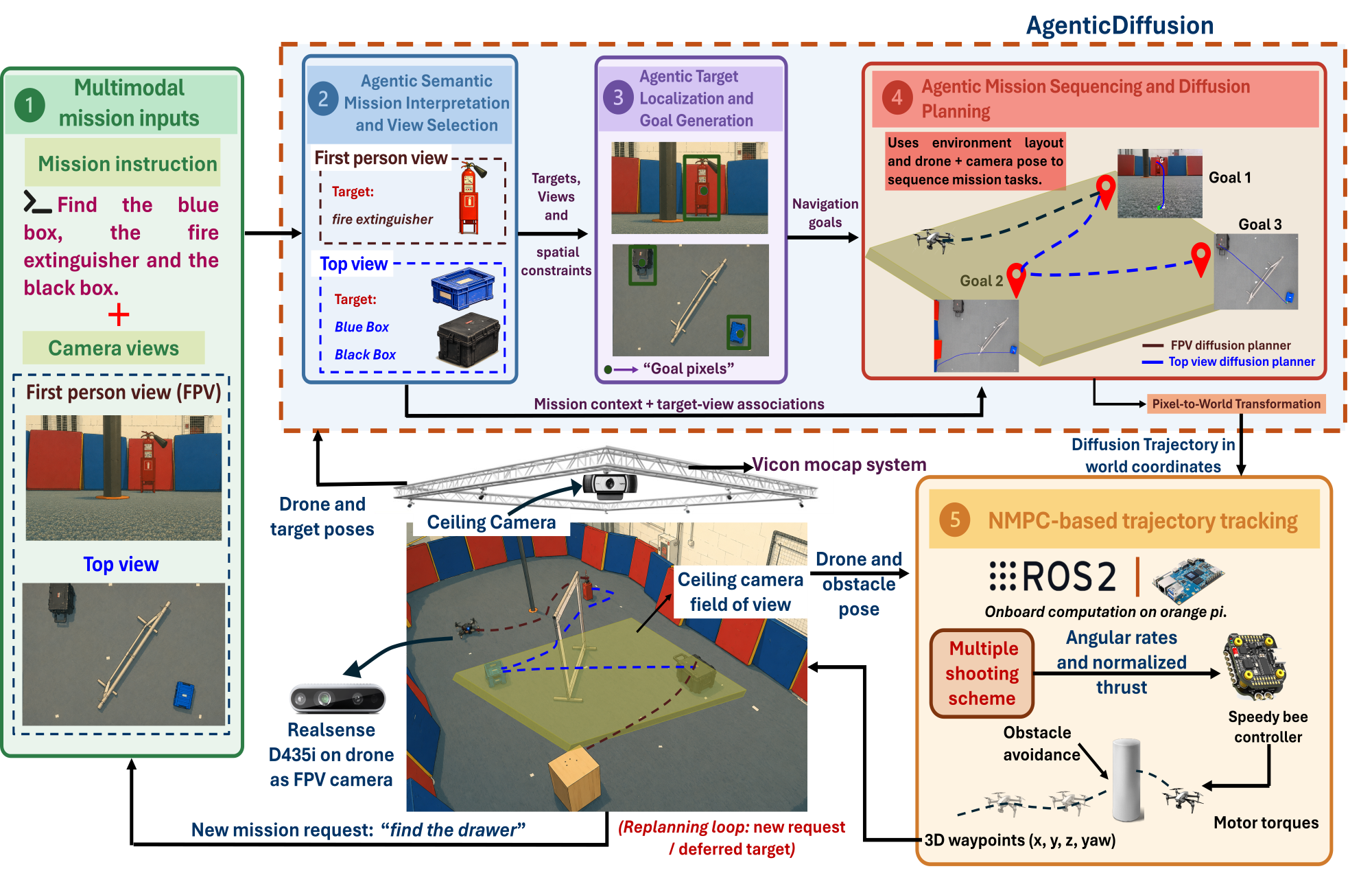}
\caption{Overview of the proposed AgenticDiffusion framework. Given a natural-language instruction together with FPV and top-view observations, the framework performs semantic mission interpretation, viewpoint selection, and target localization. The resulting navigation goal is assigned to the corresponding view-conditioned diffusion planner and executed by the UAV using NMPC. When the instruction specifies an approach direction or preferred arrival region relative to the target, the framework can instead select a suitable nearby navigation goal that satisfies the requested spatial constraint. The replanning loop is invoked for new user-issued instructions and when a previously deferred navigation task is resumed during mission continuation}
\label{fig:system_architeture}
\end{figure*}

\section{Methodology}

The proposed \textbf{AgenticDiffusion} framework follows a hierarchical pipeline comprising semantic mission reasoning, navigation-goal generation, view-conditioned trajectory planning, and NMPC-based execution, as illustrated in Fig.~\ref{fig}. AgenticDiffusion uses a semantic VLM for mission interpretation and multi-view reasoning, including view selection and target localization. Claude Sonnet 5 is used as the VLM implementation in all reported experiments ~\cite{claudesonnet}. OpenClaw serve as the agentic orchestration layer that coordinates these components and routes each navigation task to the appropriate diffusion planner~\cite{openclaw}.

\subsection{Semantic Mission Interpretation and View Selection}

Given a natural-language mission instruction, the agent identifies the requested semantic targets and associates each target with either the FPV or top-view observation. View selection considers not only target visibility but also whether the surrounding visual context is suitable for navigation.
For example, for the instruction \textit{``find the drawer and the blue box''}, the drawer is selected from the FPV observation because it is nearby and its local surroundings are clearly represented from the onboard view. In contrast, the blue box is assigned to the top view, which provides broader spatial context for navigation. The selected camera view determines the corresponding view-conditioned diffusion planner.
When a mission contains multiple targets, the resulting target--view associations are retained for subsequent mission sequencing and trajectory generation.

\subsection{Semantic Target Localization and Navigation-Goal Generation}

Once a camera view has been selected, the agent localizes the corresponding target in image space. The localized target is then converted into a navigation goal using the image-to-world transformation described in the following subsection. In the real-world indoor experiments, the targets were located below the UAV flight altitude, so the localized target position could be used directly for navigation. In the Gazebo environment, targets could lie directly along the UAV's flight path, and therefore a nearby offset goal was used to avoid navigating into the target itself.
For instructions that specify a spatial relation to a target, AgenticDiffusion supports \textit{semantic goal generation}, in which the agent selects a navigation goal that satisfies the requested relation rather than navigating directly to the localized target. For example, an instruction may require the UAV to approach the right side of a building or the rear of a structure while avoiding nearby obstacles. This mechanism is used in the Gazebo Small City experiment described in Section~\ref{sec:small_city}.

\subsection{Agentic Mission Planning}

After target localization and navigation-goal generation, the agent constructs an executable mission by sequencing the requested navigation tasks and assigning each task to the corresponding view-conditioned diffusion planner. This decision is based on the current state of the UAV, the selected camera views, the generated navigation goals, and the spatial context of the scene.

Therefore, each navigation task includes the target identity, navigation goal, selected camera view, associated diffusion planner, and position in the execution sequence. During mission execution, the agent continuously monitors the UAV state to determine when the current task has been completed and when the system should transition to the next task. If a requested target cannot be localized from the available observations, the agent can continue the search using a subsequent observation and repeat the reasoning process. When multiple navigation tasks are identified, the agent determines their execution order using the spatial arrangement of the environment, the UAV's current position, and the selected camera views. Targets located earlier along the UAV's planned direction of travel are prioritized over targets farther along the route. This geometry-based ordering rule is applied in both the real-world and Gazebo experiments.

\subsection{Vision-Based Diffusion Trajectory Planning}

AgenticDiffusion uses separate image-conditioned diffusion planners for FPV and top-view navigation. Given an RGB observation $I$, a start pixel $\mathbf{p}_s$, and a navigation-goal pixel $\mathbf{p}_g$, the selected planner generates a pixel-space trajectory connecting the start and goal locations. The top-view planner is used for longer-horizon navigation from a global scene perspective, whereas the FPV planner generates shorter-horizon egocentric trajectories. Both planners follow the same Denoising Diffusion Probabilistic Model (DDPM) formulation and are trained entirely in simulation before deployment in the real-world experiments.
The RGB observation and start--goal coordinates are encoded independently and fused into a conditioning vector:
\begin{equation}
\mathbf{c}
=
f_{\mathrm{fuse}}
\left(
f_I(I),
f_s(\mathbf{p}_s),
f_g(\mathbf{p}_g)
\right),
\label{eq:diff_cond}
\end{equation}
where $f_I$ denotes the RGB image encoder, $f_s$ and $f_g$ denote the
start- and goal-coordinate encoders, respectively, and
$f_{\mathrm{fuse}}$ combines these representations into the global
conditioning vector $\mathbf{c}$. A ResNet-18 backbone is used for image
encoding, while the pixel coordinates $\mathbf{p}_s$ and $\mathbf{p}_g$
are normalized to $[0,1]^2$ before encoding.

The diffusion state is represented as a three-channel image
\begin{equation}
x_0=[S,G,T],
\label{eq:diff_rep}
\end{equation}
where $S$, $G$, and $T$ denote the start, goal, and trajectory masks,
respectively. During training, Gaussian noise $\epsilon$ is added according to
\begin{equation}
x_t =
\sqrt{\bar{\alpha}_t}x_0
+
\sqrt{1-\bar{\alpha}_t}\,\epsilon,
\qquad
\epsilon\sim\mathcal{N}(0,\mathbf{I}),
\label{eq:forward_diff}
\end{equation}
where $x_t$ is the noisy representation at diffusion step $t$ and
$\bar{\alpha}_t=\prod_{s=1}^{t}(1-\beta_s)$ is the cumulative noise
schedule defined by the DDPM variance parameters $\beta_s$.
Rather than predicting the added noise, the conditional UNet directly
estimates the clean trajectory representation:
\begin{equation}
\hat{x}_0
=
f_{\theta}(x_t,t,\mathbf{c}),
\label{eq:x0_pred}
\end{equation}
where $f_{\theta}$ denotes the conditional UNet and $\hat{x}_0$ is the
predicted clean start--goal--trajectory representation. During reverse
diffusion, the known start and goal locations are preserved at each
denoising step so that the generated trajectory remains constrained to the
requested endpoints.
Training minimizes a weighted reconstruction objective over the trajectory
and endpoint channels:
\begin{equation}
\mathcal{L}_{\mathrm{diff}} =
\lambda_T\mathcal{L}_T
+
\frac{1}{2}
\left(
\lambda_S\mathcal{L}_S
+
\lambda_G\mathcal{L}_G
\right),
\label{eq:diff_loss}
\end{equation}
where $\mathcal{L}_T$, $\mathcal{L}_S$, and $\mathcal{L}_G$ are the
mean-squared reconstruction errors for the trajectory, start, and goal masks,
respectively, and $\lambda_T$, $\lambda_S$, and $\lambda_G$ control their
relative contributions.

\subsection{Trajectory Conversion and Reference Generation}

The trajectory produced by the selected diffusion planner is represented in image space and is subsequently transformed into a world-frame reference trajectory for control. For FPV navigation, the full predicted pixel-space trajectory is projected into the world frame using the depth image aligned with the FPV RGB observation, together with the camera intrinsic parameters, the camera-to-UAV extrinsic transformation, and the current UAV pose. For top-view navigation, the full pixel-space trajectory is converted into world coordinates using geometric projection based on the known top-view camera configuration and arena geometry.
The resulting world-frame trajectory is then sampled to form a reference sequence for tracking. 

\subsection{NMPC-Based Trajectory Execution}

The world-frame reference trajectory generated from the diffusion output is
tracked using nonlinear model predictive control (NMPC). The prediction model
uses a continuous-time 13-state quadrotor model,
\begin{equation}
\begin{aligned}
\mathbf{x}
&=
[\mathbf{p}_W^T,\mathbf{v}_W^T,\mathbf{q}_B^T,
\boldsymbol{\omega}_B^T]^T,\\
\dot{\mathbf{x}}
&=
\begin{bmatrix}
\mathbf{v}_W\\
R(q)\mathbf{T}_B/m+\mathbf{g}\\
\frac{1}{2}\mathbf{q}_{\omega B}\cdot\mathbf{q}_B\\
J^{-1}\left(\boldsymbol{\tau}_B-
\boldsymbol{\omega}_B\times J\boldsymbol{\omega}_B\right)
\end{bmatrix}.
\end{aligned}
\label{eq:quad_dynamics}
\end{equation}
where $\mathbf{p}_W=[p_x,p_y,p_z]^T$ and
$\mathbf{v}_W=[v_x,v_y,v_z]^T$ denote the world-frame position and
velocity, $\mathbf{q}_B=[q_w,q_x,q_y,q_z]^T$ is the attitude quaternion,
and $\boldsymbol{\omega}_B=[\omega_x,\omega_y,\omega_z]^T$ contains the
body angular rates. Here,
$\mathbf{T}_B=[0,0,\sum_{i=1}^{4}T_i]^T$ is the total thrust,
$\mathbf{g}=[0,0,9.81]^T$ is the gravitational vector,
$J=\mathrm{diag}(J_x,J_y,J_z)$ is the inertia matrix, and
$\mathbf{q}_{\omega}=(0,\omega_x,\omega_y,\omega_z)^T$ is the angular
velocity quaternion.

The body torque is given by
\begin{equation}
\boldsymbol{\tau}_B=
\begin{bmatrix}
-l_y & l_y & l_y & -l_y\\
-l_x & -l_x & l_x & l_x\\
k_t & -k_t & k_t & -k_t
\end{bmatrix}
\begin{bmatrix}
u_1\\u_2\\u_3\\u_4
\end{bmatrix},
\label{eq:torque_matrix}
\end{equation}
where $u_1,\ldots,u_4$ are the motor forces, $l_x$ and $l_y$ are the
distances to the $x$- and $y$-axes, respectively, and $k_t$ is the torque
constant.

At each control step, NMPC solves the finite-horizon optimization problem
\begin{equation}
\begin{aligned}
\min_{\mathbf{x}_{0:N},\mathbf{u}_{0:N-1}}
\quad &
\sum_{i=0}^{N-1}
\left\|
\mathbf{y}_i-\mathbf{y}^{\mathrm{ref}}_i
\right\|_{W}^{2}
+
\left\|
\mathbf{y}_N-\mathbf{y}^{\mathrm{ref}}_N
\right\|_{W_N}^{2}
\\
\mathrm{s.t.}\quad &
\mathbf{x}_0=\mathbf{x}_{\mathrm{meas}},\qquad
\mathbf{x}_{i+1}
=
f_{\mathrm{RK4}}(\mathbf{x}_i,\mathbf{u}_i),
\\
&
\mathbf{x}_{\min}\leq\mathbf{x}_i\leq\mathbf{x}_{\max},\qquad
\mathbf{u}_{\min}\leq\mathbf{u}_i\leq\mathbf{u}_{\max}.
\end{aligned}
\label{eq:nmpc_compact}
\end{equation}
where $N$ is the prediction horizon,
$\mathbf{x}_{\mathrm{meas}}$ is the current measured state, and
$f_{\mathrm{RK4}}$ denotes the discrete-time dynamics obtained using
fourth-order Runge--Kutta integration. The vectors $\mathbf{y}_i$ and
$\mathbf{y}^{\mathrm{ref}}_i$ denote the predicted and reference outputs,
while $W$ and $W_N$ are the stage and terminal weighting matrices.
Only the first optimized control input is applied before the optimization is
repeated using the updated UAV state, enabling receding-horizon tracking of
the diffusion-generated trajectory.

\section{Experimental Setup and Results}
AgenticDiffusion was evaluated in four real-world indoor UAV navigation scenarios and an additional experiment in the simulated Gazebo Small City environment. Each real-world scenario was repeated 10 times, yielding 40 physical-flight trials. The real-world experiments assessed multi-target navigation, semantic view selection, and sequential mission execution using complementary FPV and top-view observations, while the Gazebo experiment examined the framework under a different visual and spatial configuration.

The real-world experiments were conducted using a custom-built quadrotor equipped with a SpeedyBee flight controller, an Orange Pi 5B onboard computer, and an Intel RealSense D435i RGB-D camera providing the FPV observation at 15~Hz. A Logitech C930ce camera provided the external top-view observation. The two cameras were spatially separated by more than 1.8~m to provide complementary coverage of the experimental arena. In the Gazebo Small City experiment, both the FPV and top-view cameras provided RGB-D observations.
The agentic orchestration and diffusion-planning components were hosted on a desktop PC equipped with an NVIDIA RTX 5090 GPU with 24~GB of available GPU memory. Communication between the planning and control components was implemented using ROS~2, while the NMPC controller operated at 100~Hz.
\subsection{Scenario 1: Multi-Stage Multi-View Semantic Navigation}

\begin{figure}[htbp]
\centering
\includegraphics[width=1\linewidth]{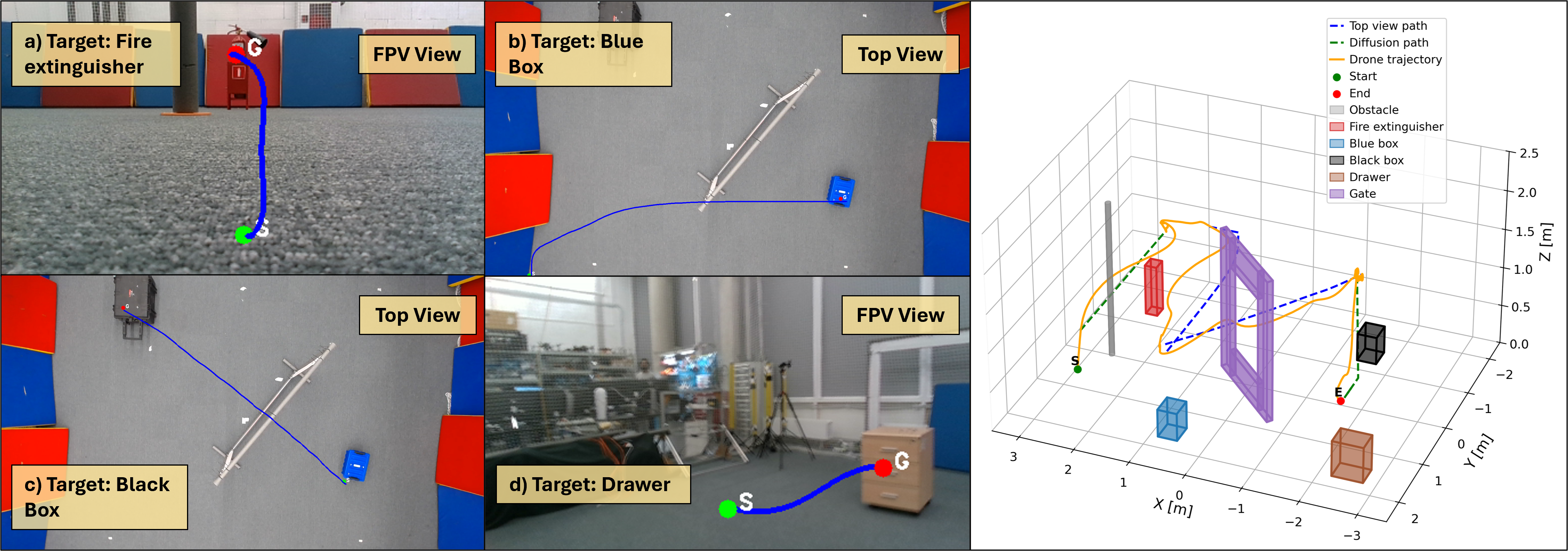}
\caption{Scenario 1: Sequential multi-view navigation toward the fire extinguisher, blue box, black box, and drawer using FPV and top-view diffusion planning.}
\label{fig}
\end{figure}

Scenario 1 evaluated sequential navigation to multiple targets using complementary camera views, as shown in Fig.~\ref{fig}. For the instruction \textit{``find the blue box, fire extinguisher, and the black box,''} the agent assigned the fire extinguisher to FPV navigation and the blue and black boxes to top-view navigation. Based on the geometry-based mission-ordering rule, the fire extinguisher was selected as the first target because it was encountered earlier along the UAV's progression through the arena. The mission then continued with top-view navigation toward the blue box and black box.

A subsequent instruction requested the drawer, which was assigned to the FPV planner based on its representation in the FPV observation. The framework completed 8 of 10 trials successfully (80\%); the two unsuccessful trials resulted from communication loss between the VICON system and the UAV.

\subsection{Scenario 2: Multi-View Target Selection}

Scenario 2 examined view selection when a requested target was visible from more than one camera perspective, as shown in Fig.~\ref{fig}. For the instruction \textit{``find the drawer and the blue box,''} the drawer was visible in both FPV and top-view observations, but the agent selected the FPV view because it provided a more suitable local representation for navigation. The blue box was assigned to the top-view planner. The resulting mission therefore executed FPV navigation toward the drawer followed by top-view navigation toward the blue box.

The framework completed 7 of 10 trials successfully (70\%). Failures occurred in highly cluttered configurations where the NMPC obstacle-avoidance cost caused the UAV to hover near an obstacle rather than deviate sufficiently from the reference trajectory. Larger avoidance maneuvers were limited by the VICON tracking region, as leaving the capture volume could result in localization loss.

\begin{figure}[htbp]
\centering
\includegraphics[width=1\linewidth]{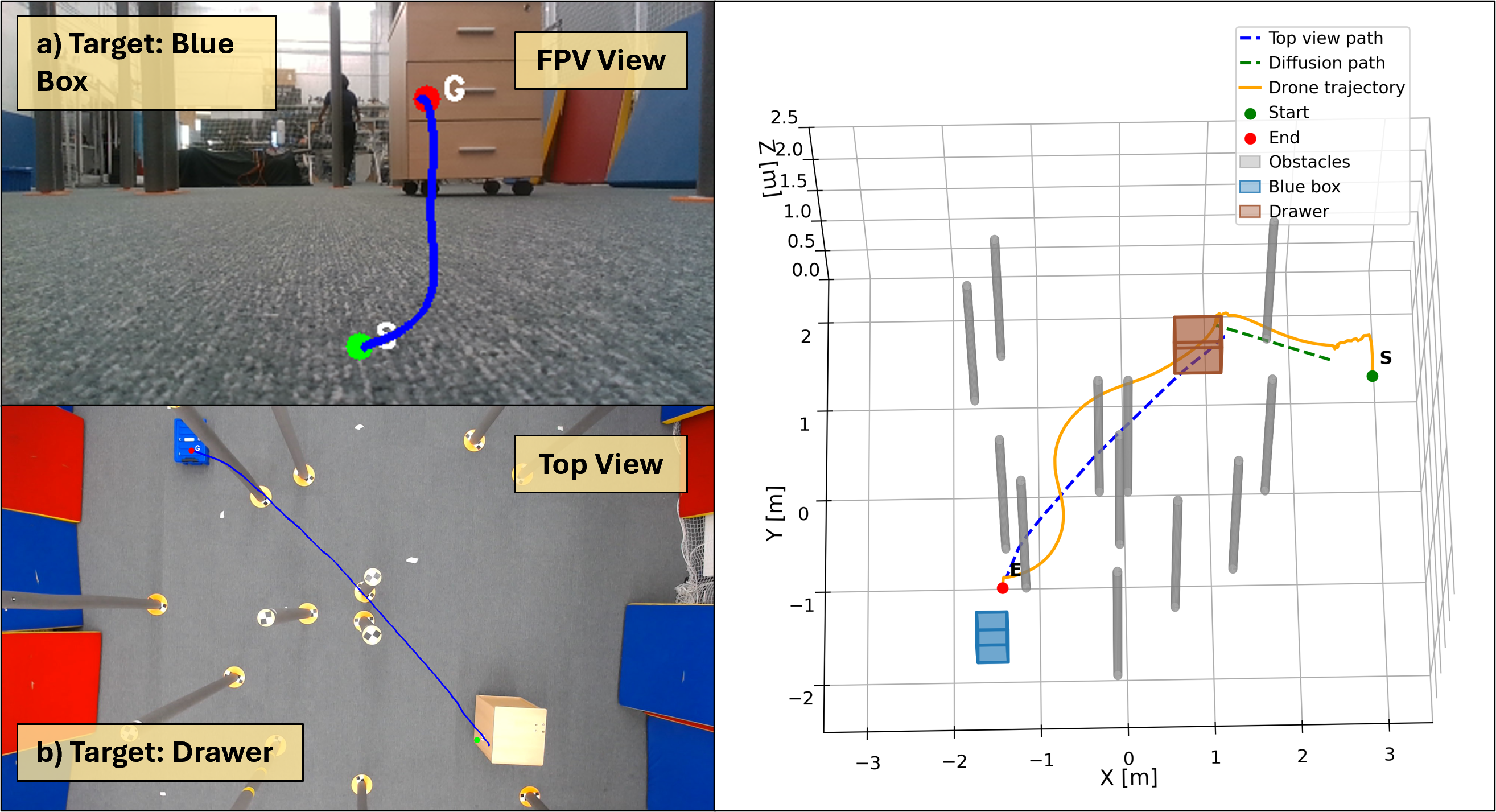}
\caption{Scenario 2: Multi-view target selection when the drawer is visible in both camera views but is assigned to FPV navigation, while the blue box is assigned to the top-view planner.}
\label{fig}
\end{figure}

\subsection{Scenario 3: Long-Horizon Multi-Target Navigation}

\begin{figure}[t]
\centering
\includegraphics[width=1\linewidth]{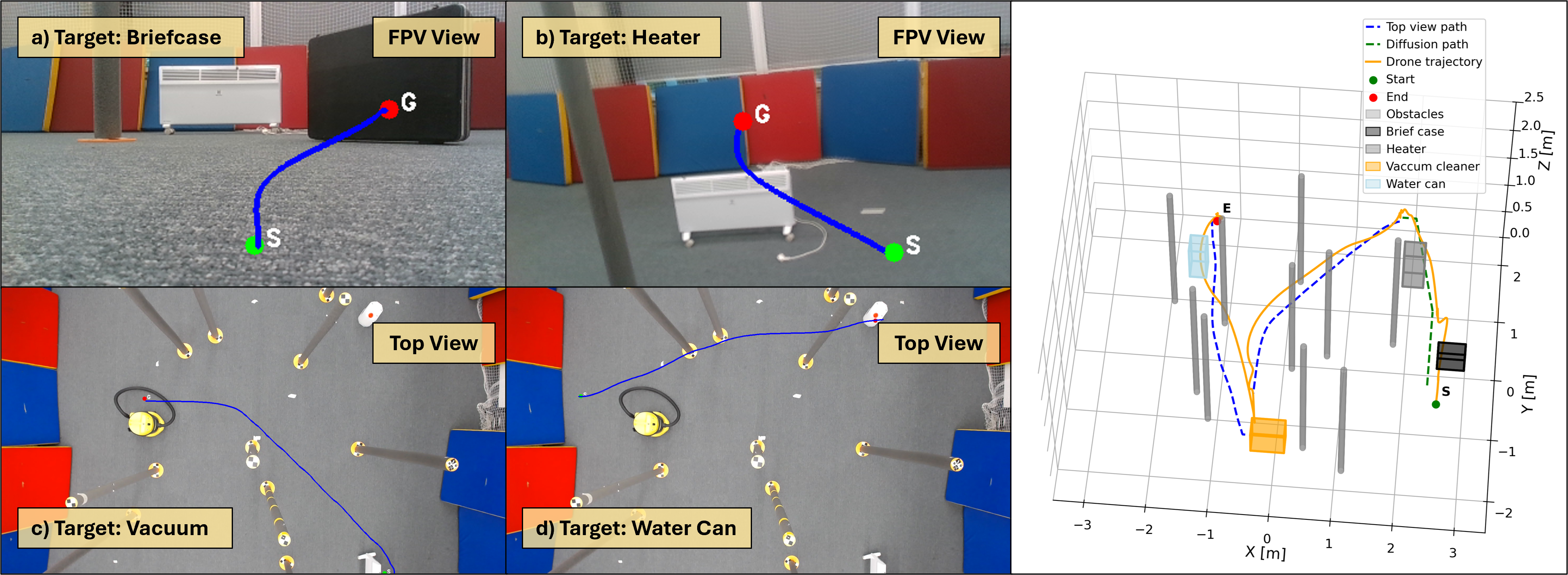}
\caption{Scenario 3: Long-horizon multi-target navigation using sequential FPV and top-view planning.}
\label{fig:exp3}
\end{figure}

Scenario 3 evaluated multi-target navigation over a longer mission sequence, as illustrated in Fig.~\ref{fig:exp3}. For the instruction \textit{``find the briefcase, vacuum cleaner, heater, and the water can,''} the agent assigned the briefcase and heater to FPV navigation and the vacuum cleaner and water can to top-view navigation. Following the mission-ordering strategy described in the Methodology, the FPV tasks were executed first, followed by the top-view tasks.
The framework completed 7 of 10 trials successfully (70\%). The unsuccessful trials resulted from NMPC local deadlock near obstacles.

\subsection{Scenario 4: Semantic Landing-Site Selection}

\begin{figure}[htbp]
\centering
\includegraphics[width=1\linewidth]{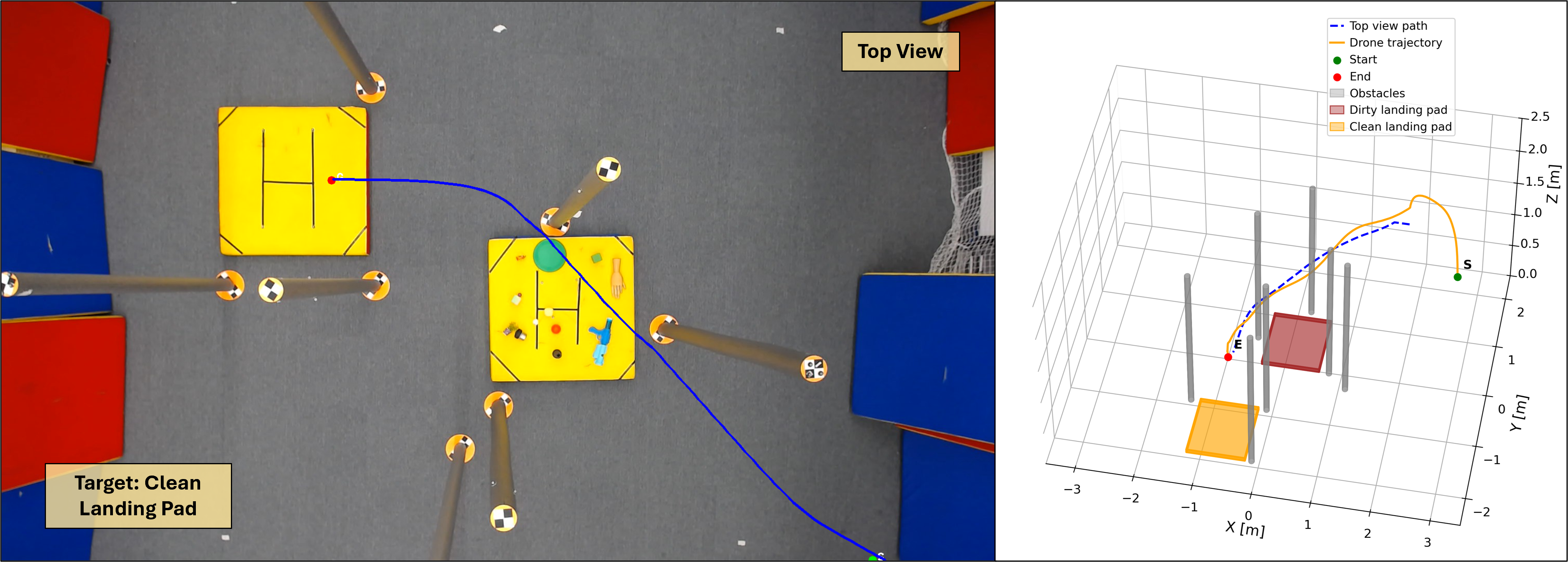}
\caption{Scenario 4: Semantic landing-site selection using multi-view observations and surrounding obstacle context.}
\label{fig:exp4}
\end{figure}

Scenario 4 examined whether the agent could use scene context to select between two candidate landing sites, as shown in Fig.~\ref{fig:exp4}. Given the instruction \textit{``find the cleanest landing site for the drone,''} the agent compared the surrounding obstacle distributions of the two candidates. Although both were partially visible in the FPV observation, the top view provided clearer spatial context, and the agent selected the less cluttered site for navigation.
The selected landing site was assigned to the top-view diffusion planner, and the framework completed all 10 trials successfully (100\%).

\begin{figure*}[t]
\centering
\includegraphics[width=1\linewidth]
{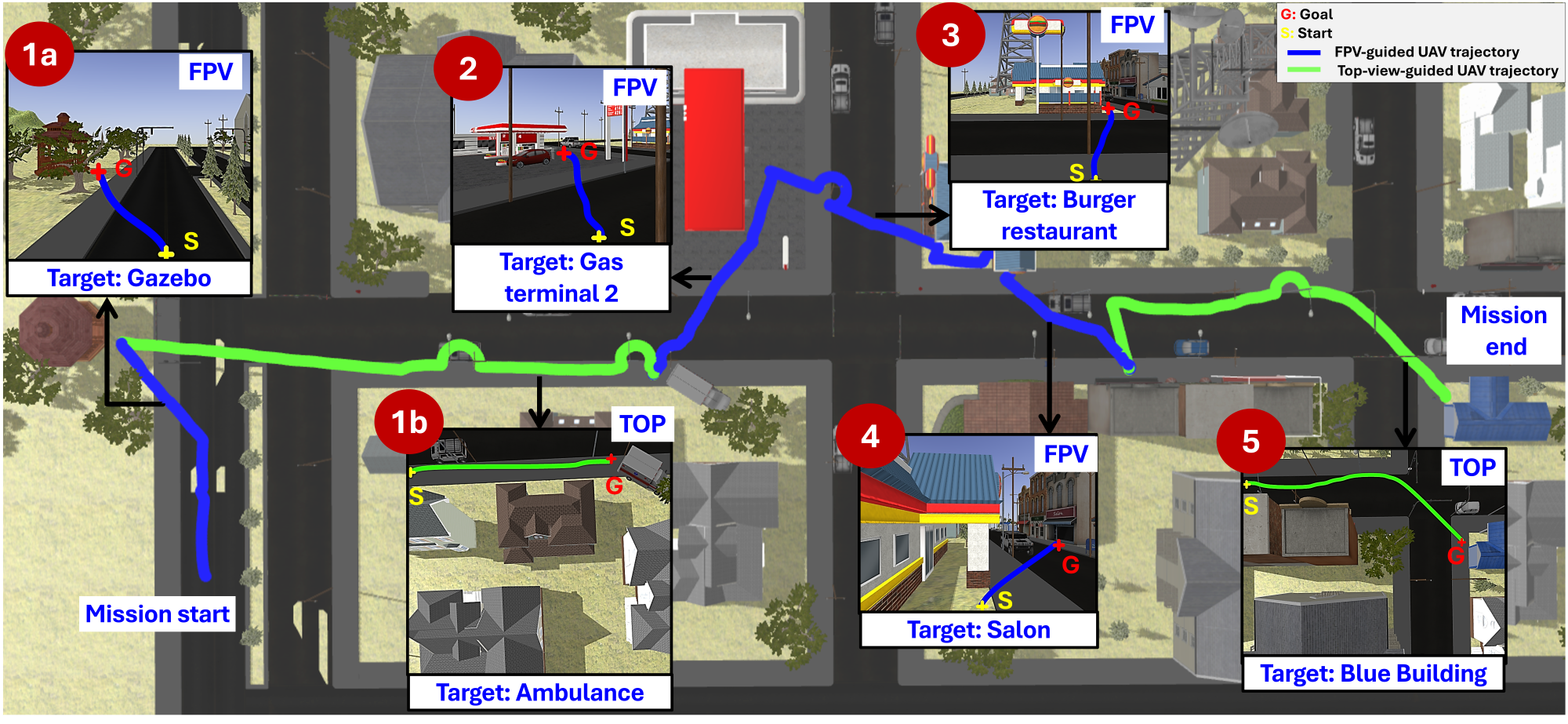}
\caption{Gazebo Small City evaluation of AgenticDiffusion. Drone~1 provides FPV observations and executes the planned trajectories, while Drone~2 provides complementary top-view observations from predefined non-overlapping coverage locations.}
\label{fig:gazebo_simulation}
\end{figure*}

\subsection{Scenario 5: Multi-View Navigation in Gazebo Small City}
\label{sec:small_city}
Scenario~5 evaluates AgenticDiffusion in the Gazebo Small City environment,
which contains roads, buildings, and vehicles and requires substantially longer
navigation distances than the indoor experiments. Drone~1 provides FPV
observations and executes the generated trajectories, while Drone~2 provides
external top-view observations from predefined locations across the environment.
Drone~2 follows a fixed coverage pattern that progressively exposes new regions
while minimizing overlap between successive top-view observations. This provides
complementary scene coverage as the mission progresses.
The overall mission consists of five sequential mission instructions:

\begin{enumerate}
\item \textbf{Instruction 1:}
\textit{``Find the gazebo and the ambulance.''}

\item \textbf{Instruction 2:}
\textit{``Find the burger restaurant from the right side of the road, then go to
Terminal 2 from the rear side of the gas station and avoid all cars.''}

\item \textbf{Instruction 3:}
\textit{``Go to the burger restaurant from the right side of the road. Select a
landing point on the flat road to the right of the restaurant in the image.''}

\item \textbf{Instruction 4:}
\textit{``Go to the salon.''}

\item \textbf{Instruction 5:}
\textit{``Go to the blue building.''}
\end{enumerate}

Instructions~1, 2, 4, and~5 are issued by the user, whereas Instruction~3 is
generated autonomously by the agent as a continuation of Instruction~2 after
the burger restaurant is identified but temporarily deferred while the UAV
navigates to Terminal~2 from the rear side of the gas station.

For each requested target, the agent selects between FPV and top-view navigation
based on the available visual context. In Instruction~1, for example, the gazebo
is assigned to the FPV diffusion planner, whereas the ambulance is assigned to
the top-view diffusion planner. Instructions~2 and~3 additionally require
spatially constrained navigation. In these cases, semantic goal generation
allows the agent to distinguish the localized target from the desired navigation
goal and select a nearby point that satisfies the requested spatial relation.

Across repeated Gazebo runs, AgenticDiffusion achieved a \textbf{90\% mission success rate}. The remaining failures were associated with transient depth spikes during navigation-goal generation, which occasionally displaced the final waypoint from the intended region. These failures arose from depth-based geometric goal refinement rather than incorrect semantic view selection.

\section{Comparative Analysis}
The comparative analysis addresses two research questions:
\textbf{RQ1} investigates multi-view navigation efficiency and mission completion relative to single-view navigation, while \textbf{RQ2} evaluates the robustness of semantic view selection under target-description variation.

\subsection{RQ1: Efficiency and Success of Multi-View Navigation}
RQ1 examines whether coordinating FPV and top-view observations reduces mission time and supports completion of navigation tasks compared with FPV-only and Top-only navigation. \textit{Camera Views} denotes the total number of observations used during a mission, while a \textit{Target-Search Action} denotes the acquisition of an additional observation when a requested target cannot be resolved from the current view.
For FPV-only navigation, each additional search action requires approximately a $30^\circ$ UAV rotation followed by target inspection through the semantic reasoning stage, resulting in an estimated overhead of 15~s. For Top-only navigation, the external camera is repositioned in the $x$--$y$ plane before a new observation is acquired and evaluated, resulting in an estimated overhead of 30~s. AgenticDiffusion mission times are averaged over 10 executions, with human prompting time excluded.
\begin{table}[t]
\centering
\caption{Comparison of multi-view and single-view navigation.}
\label{tab:rq1_multiview}
\footnotesize
\setlength{\tabcolsep}{2.3pt}
\renewcommand{\arraystretch}{1.05}
\resizebox{\columnwidth}{!}{
\begin{tabular}{llcccc}
\hline
\textbf{Scenario} &
\textbf{Method} &
\shortstack{\textbf{Camera}\\\textbf{Views}} &
\shortstack{\textbf{Target-Search}\\\textbf{Actions}} &
\shortstack{\textbf{Avg. Time}\\\textbf{(s)}} &
\shortstack{\textbf{Mission} \\\textbf{Complete}} \\
\hline

S1 & AgenticDiffusion & 3 & 1 & 61.2 & Yes \\
   & FPV-only         & 7 & 3 & 135.0 & Yes \\
   & Top-only         & 2 & 1 & 80.4 & Yes \\
\hline

S2 & AgenticDiffusion & 2 & 0 & 47.2 & Yes \\
   & FPV-only         & 3 & 1 & 65.3 & Yes \\
   & Top-only         & 1 & 0 & 45.0 & No \\
\hline

S3 & AgenticDiffusion & 2 & 0 & 73.0 & Yes \\
   & FPV-only         & 5 & 2 & 137.9 & Yes \\
   & Top-only         & 2 & 1 & 102.9 & Yes \\
\hline

S4 & AgenticDiffusion & 1 & 0 & 21.0 & Yes \\
   & FPV-only         & -- & -- & -- & No \\
   & Top-only         & 1 & 0 & 21.0 & Yes \\
\hline

Gazebo & AgenticDiffusion & 10 & 1 & 228.0 & Yes \\
       & FPV-only         & 11 & 5 & 280.0 & No \\
       & Top-only         & 9 & 5 & 258.4 & No \\
\hline
\end{tabular}}
\end{table}
As shown in Table~\ref{tab:rq1_multiview}, the largest efficiency gain occurs in Scenarios~1 and~3, where targets are distributed across complementary viewpoints. Across these mixed-visibility scenarios, AgenticDiffusion reduced average mission time by \textbf{50.8\%} relative to FPV-only navigation and by \textbf{26.8\%} relative to Top-only navigation.
The results also show that shorter execution time does not necessarily imply successful mission completion. In Scenario~2, the Top-only variant could not reliably complete the drawer task because the drawer was not detected consistently from the top view when its description was varied. In Scenario~4, the FPV-only variant lacked sufficient surrounding spatial context to compare the clutter around the candidate landing sites and select the more suitable one. In Gazebo, neither single-view variant completed the full multi-stage mission: FPV-only could not resolve the blue building, whereas Top-only could not complete intermediate targets including Gas station terminal~2 and the salon.

\subsection{RQ2: Robustness of Semantic View Selection}

RQ2 evaluates whether semantic view selection remains robust when the same physical target is expressed using different textual descriptions. Eleven target instances from Scenarios~1--4 were evaluated using one canonical description and five alternative descriptions per target, yielding 66 cases. For each target, a ground-truth preferred view was predefined based on target visibility and the surrounding spatial context required for navigation.

AgenticDiffusion uses a semantic VLM for view selection, instantiated using Claude Sonnet~5 in the reported experiments. As a baseline, we compare this with a confidence-based Grounding DINO selector. The semantic VLM jointly considers the paired FPV and top-view observations and selects the preferred navigation view, whereas Grounding DINO processes the observations independently and selects the view with the higher detection confidence. Performance is evaluated using \textit{view-selection accuracy} and \textit{lexical consistency},where lexical consistency measures whether alternative descriptions preserve the view selected for the canonical description.

As shown in Table~\ref{tab:rq2_view}, the semantic VLM preserved the ground-truth preferred view across all 66 cases, whereas the confidence-based Grounding DINO baseline produced 14 view changes among the 55 alternative-description cases. This indicates that detection confidence alone
does not consistently reflect which observation provides the more suitable navigation context.
Localization results provide additional insight into this behavior. For canonical object descriptions, the semantic VLM and Grounding DINO produced comparable localization performance, with mean intersection over union (IoU)s of 0.6725 and 0.7073, respectively. Under alternative descriptions, however, the semantic VLM maintained a higher mean IoU (0.6941 versus 0.6643) and a lower overall normalized center error (0.0123 versus 0.0361). Both methods returned bounding boxes in all 66 cases. These results suggest that the main advantage of semantic reasoning emerges under lexical variation and context-dependent target interpretation rather than conventional object localization alone. This capability is also relevant to the Gazebo scenario, where navigation instructions include spatially constrained target descriptions. Scenario~2 shows the same behavior: the drawer was reliably identified in the FPV view but not consistently from the top view under alternative descriptions.

\begin{table}[t]
\centering
\caption{RQ2 view-selection robustness under target-description variation.}
\label{tab:rq2_view}
\footnotesize
\renewcommand{\arraystretch}{1.05}

\begin{tabular*}{\columnwidth}{@{\extracolsep{\fill}}lccc@{}}
\hline
\textbf{Method} &
\shortstack{\textbf{View-Selection}\\\textbf{Accuracy}} &
\shortstack{\textbf{Lexical}\\\textbf{Consistency}} &
\shortstack{\textbf{View}\\\textbf{Flips}} \\
\hline
Semantic VLM   & 100.00\% & 100.00\% & 0  \\
Grounding DINO & 63.64\%  & 74.55\%  & 14 \\
\hline
\end{tabular*}
\end{table}

\vspace{-4pt}
\section{Discussion and Limitations}
The results highlight the importance of complementary viewpoints for semantic UAV navigation. The largest efficiency gains were observed when requested targets were distributed across FPV and top-view observations, where restricting navigation to a single viewpoint introduced additional search actions or prevented completion of the full mission. This behavior was also observed in the Gazebo Small City experiment: AgenticDiffusion achieved a 90\% mission success rate and completed the multi-stage mission, whereas the FPV-only and Top-only variants could not resolve all requested targets. The view-selection experiments further showed that semantic reasoning remained consistent under lexical variation, while confidence-based selection was more sensitive to changes in target description.

The current framework relies on known approximate environment geometry and predefined operating regions for the FPV and top-view platforms. The real-world implementation additionally uses VICON for UAV localization, obstacle information, and mission-state monitoring within a $3\times3$~m arena, which restricts the scale of physical evaluation. Target search is currently restricted to local observation changes, including UAV rotation for FPV search and predefined repositioning or coverage patterns for the external viewpoint, rather than active exploration. 

\section{Conclusion and Future Work}
\label{sec:conclusion}
This paper presented \textbf{AgenticDiffusion}, an agentic multi-view UAV navigation framework that coordinates FPV and top-view observations for semantic mission planning, view-conditioned diffusion trajectory generation, and NMPC-based execution. Across 40 real-world flight trials, the framework achieved an overall mission success rate of \textbf{80\%}. In mixed-visibility scenarios, coordinated multi-view navigation reduced average mission time by \textbf{50.8\%} relative to FPV-only navigation and by \textbf{26.8\%} relative to Top-only navigation. Semantic view selection achieved \textbf{100\% accuracy across 66 target-description cases}, compared with \textbf{63.64\%} for the confidence-based Grounding DINO baseline. In the Gazebo Small City evaluation, AgenticDiffusion achieved a \textbf{90\% mission success rate} and completed the multi-stage mission using complementary viewpoints, whereas neither single-view variant could complete all requested navigation tasks.

Future work will investigate active viewpoint acquisition and next-best-view selection in place of the current predefined search strategies. Additional directions include reducing dependence on external motion-capture infrastructure through onboard localization and mapping, improving robustness to depth uncertainty, incorporating closed-loop replanning, and evaluating the framework across a broader range of environments, mission lengths, spatial instructions, and dynamic obstacles.

\bibliographystyle{IEEEtran}
\bibliography{IEEEexample}

@ARTICLE{navrl,
  author={Xu, Zhefan and Han, Xinming and Shen, Haoyu and Jin, Hanyu and Shimada, Kenji},
  journal={IEEE Robotics and Automation Letters}, 
  title={{NavRL: Learning Safe Flight in Dynamic Environments}}, 
  year={2025},
  number={4},
  pages={3668--3675},
  volume={10},
  month = {Feb, 26,},
  
  
 }

@misc{3drldwa,
      title={{3D RL-DWA: A Hybrid Reinforcement Learning and Dynamic Window Approach for Goal-Directed Local Navigation in Multi-DoF Robots}}, 
      author={Chiara Castellani and Enrico Turco and Domenico Prattichizzo},
      year={2026},
      note={arxiv:2605.12689}, 
}

@misc{macroactions,
      title={{Deep Reinforcement Learning Based Navigation with Macro Actions and Topological Maps}}, 
      author={Simon Hakenes and Tobias Glasmachers},
      year={2025},
      eprint={2504.18300},
      archivePrefix={arXiv},
      primaryClass={cs.LG},
     note={arxiv:2504.18300}, 
}

@INPROCEEDINGS{dipper,
  author={Liu, Jianwei and Stamatopoulou, Maria and Kanoulas, Dimitrios},
  booktitle={Proc. IEEE Int. Conf. on Robotics and Automation (ICRA)}, 
  title={{DiPPeR: Diffusion-based 2D Path Planner applied on Legged Robots}}, 
  month = {May 13-17,},
  year={2024},
  volume={},
  number={},
  pages={9264--9270}}

@INPROCEEDINGS{dtg,
  author={Liang, Jing and Payandeh, Amirreza and Song, Daeun and Xiao, Xuesu and Manocha, Dinesh},
  booktitle={Proc. IEEE Int. Conf. on Intelligent Robots and Systems (IROS)}, 
  title={{DTG: Diffusion-based Trajectory Generation for Mapless Global Navigation}}, 
  year={2024},
  volume={},
  number={},
month = {Oct. 14-18,},
  pages={5340--5347}}

@INPROCEEDINGS{nomad,
  author={Sridhar, Ajay and Shah, Dhruv and Glossop, Catherine and Levine, Sergey},
  booktitle={Proc. IEEE Int. Conf. on Robotics and Automation (ICRA)}, 
  title={{NoMaD: Goal Masked Diffusion Policies for Navigation and Exploration}}, 
  year={2024},
  volume={},
  number={},
  month = {May 13-17,},
  pages={63--70},}

@misc{navdp,
      title={{NavDP: Learning Sim-to-Real Navigation Diffusion Policy with Privileged Information Guidance}}, 
      author={Wenzhe Cai and Jiaqi Peng and Yuqiang Yang and Yujian Zhang and Meng Wei and Hanqing Wang and Yilun Chen and Tai Wang and Jiangmiao Pang},
      year={2025},
      eprint={2505.08712},
      note={arXiv:2505.08712}, 
}

@INPROCEEDINGS{NavDiffuser,
  author={Zeng, Yiming and Ren, Hao and Wang, Shuhang and Huang, Junlong and Cheng, Hui},
  booktitle={Proc. IEEE Int. Conf. on on Robotics and Automation (ICRA)}, 
  title={{NaviDiffusor: Cost-Guided Diffusion Model for Visual Navigation}}, 
  year={2025},
  month = {May 20-23,},
  volume={},
  number={},
  pages={11994--12001}}

@inproceedings{10.5555/3507788.3507815,
author = {Liu, Xiaotian and Armstrong, Victoria and Nabil, Sara and Muise, Christian},
title = {Exploring multi-view perspectives on deep reinforcement learning agents for embodied object navigation in virtual home environments},
year = {2021},
month = {Nov. 22-25,},
booktitle = {Proc. of Int. Conf. on Computer Science and Software Engineering (CASCON)},
pages = {190–195},

}

@article{YANG2017116,
author = {Shaowu Yang and Sebastian A. Scherer and Xiaodong Yi and Andreas Zell},
title = {{Multi-camera visual SLAM for autonomous navigation of micro aerial vehicles}},
journal = {Robotics and Autonomous Systems},
volume = {93},
pages = {116--134},
year = {2017},
month = {July},
}

@INPROCEEDINGS{9341304,
  author={Zhu, Kunyan and Chen, Wei and Zhang, Wei and Song, Ran and Li, Yibin},
  booktitle={Proc. IEEE/RSJ Int. Conf. on Intelligent Robots and Systems (IROS)}, 
  title={{utonomous Robot Navigation Based on Multi-Camera Perception}}, 
  year={2020},
  volume={},
  number={},
  pages={5879--5885},
  }

@inproceedings{10.5555/3600270.3600510,
author = {Lu, Hui and Chiquier, Mia and Vondrick, Carl},
title = {Private multiparty perception for navigation},
year = {2022},
month = {Nov.-Dec. 28-09,},
booktitle = {Proc. of the Int. Conf. on Neural Information Processing Systems (NeurIPS)},
pages = {3318--3328}
}

@misc{xu2025mmnavmultiviewvlamodel,
      title={{MM-Nav: Multi-View VLA Model for Robust Visual Navigation via Multi-Expert Learning}}, 
      author={Tianyu Xu and Jiawei Chen and Jiazhao Zhang and Wenyao Zhang and Zekun Qi and Minghan Li and Zhizheng Zhang and He Wang},
      year={2025},
      note={arxiv:2510.03142},
}

@INPROCEEDINGS{huang2023visuallanguagemapsrobot,
  author={Huang, Chenguang and Mees, Oier and Zeng, Andy and Burgard, Wolfram},
  booktitle={Proc. IEEE Int. Conf. on Robotics and Automation (ICRA)}, 
  title={{Visual Language Maps for Robot Navigation}}, 
  year={2023},
  month = {May-June, 29-2,},
  volume={},
  number={},
  pages={10608--10615},}

@INPROCEEDINGS{10658310,
  author={Chen, Boyuan and Xu, Zhuo and Kirmani, Sean and Ichter, Brian and Sadigh, Dorsa and Guibas, Leonidas and Xia, Fei},
  booktitle={Proc. IEEE/CVF Conf. on Computer Vision and Pattern Recognition (CVPR)}, 
  title={{SpatialVLM: Endowing Vision-Language Models with Spatial Reasoning Capabilities}}, 
  year={2024},
  month = {June 16-22,},
  volume={},
  number={},
  pages={14455--14465},
  }

@misc{zhang2026mapnavnovelmemoryrepresentation,
      title={{MapNav: A Novel Memory Representation via Annotated Semantic Maps for Vision-and-Language Navigation}}, 
      author={Lingfeng Zhang and Xiaoshuai Hao and Qinwen Xu and Qiang Zhang and Xinyao Zhang and Pengwei Wang and Jing Zhang and Zhongyuan Wang and Shanghang Zhang and Renjing Xu},
      year={2026},
      note={arxiv:2502.13451}, 
}

@inproceedings{
lee2025imagineverifyexecutememoryguided,
title={{Imagine, Verify, Execute: Memory-guided Agentic Exploration with Vision-Language Models}},
author={Seungjae Lee and Daniel Ekpo and Haowen Liu and Furong Huang and Abhinav Shrivastava and Jia-Bin Huang},
booktitle={Conference on Robot Learning (CoRL)},
year={2025},
month = {Sept. 27-30,},
pages={4837--4858},

}

@INPROCEEDINGS{11063767,
  author={Wang, Zeyu and Yang, Xu},
  booktitle={Proc. IEEE Int. Conf. on Neural Networks, Information and Communication Engineering (NNICE)}, 
  title={{Enhancing Vision-and-Language Navigation in Continuous Environment via Data Synthesis}}, 
  year={2025},
  month = {Jan. 10-12,},
  volume={},
  number={},
  pages={713--716},
  }

@article{navid,
        title={{NaVid: Video-based VLM Plans the Next Step for Vision-and-Language Navigation}},
        author={Zhang, Jiazhao and Wang, Kunyu and Xu, Rongtao and Zhou, Gengze and Hong, Yicong and Fang, Xiaomeng and others},
        journal={Robotics: Science and Systems},
        year={2024}
      }

@misc{fastsmartway,
      title={{Fast-SmartWay: Panoramic-Free End-to-End Zero-Shot Vision-and-Language Navigation}}, 
      author={Xiangyu Shi and Zerui Li and Yanyuan Qiao and Qi Wu},
      year={2025},
      note={arxiv:2511.00933}, 
}

@inproceedings{janusvln,
title={{Janus{VLN}: Decoupling Semantics and Spatiality with Dual Implicit Memory for Vision-Language Navigation}},
author={Shuang Zeng and Dekang Qi and Xinyuan Chang and Feng Xiong and Xie Shichao and Xiaolong Wu and others},
booktitle={Int. Conf. on Learning Representations},
year={2026},
month = {April 23-25,},
}

@misc{agentvln,
      title={{AgentVLN: Towards Agentic Vision-and-Language Navigation}}, 
      author={Zihao Xin and Wentong Li and Yixuan Jiang and Ziyuan Huang and Bin Wang and Piji Li and others},
      year={2026},
      note={arxiv:2603.17670}, 
}

@misc{flowagent,
      title={{Action Agent: Agentic Video Generation Meets Flow-Constrained Diffusion}}, 
      author={Jeffrin Sam and Nguyen Khang and Yara Mahmoud and Miguel Altamirano Cabrera and Dzmitry Tsetserukou},
      year={2026},
      note={arxiv:2605.01477}, 
}

@inproceedings{openuav,
title={{Towards Realistic {UAV} Vision-Language Navigation: Platform, Benchmark, and Methodology}},
author={Xiangyu Wang and Donglin Yang and Ziqin Wang and Hohin Kwan and Jinyu Chen and Wenjun Wu and others},
booktitle={Int. Conf. on Learning Representations},
year={2025},
month = {April 24-28,},
}

@article{
wang2023voyageropenendedembodiedagent,
title={{Voyager: An Open-Ended Embodied Agent with Large Language Models}},
author={Guanzhi Wang and Yuqi Xie and Yunfan Jiang and Ajay Mandlekar and Chaowei Xiao and Yuke Zhu and Linxi Fan and Anima Anandkumar},
journal={Transactions on Machine Learning Research},
year={2024},
month = {March}

}

@INPROCEEDINGS{10802322,
  author={Kannan, Shyam Sundar and Venkatesh, Vishnunandan L. N. and Min, Byung-Cheol},
  booktitle={Proc. IEEE/RSJ Int. Conf. on Intelligent Robots and Systems (IROS)}, 
  title={{SMART-LLM: Smart Multi-Agent Robot Task Planning using Large Language Models}}, 
  year={2024},
  month = {Oct. 14-18,},
  volume={},
  number={},
  pages={12140--12147}}

@misc{herron2025hierarchicalagenticframeworkautonomous,
      title={{A Hierarchical Agentic Framework for Autonomous Drone-Based Visual Inspection}, 
      author={Ethan Herron and Xian Yeow Lee and Gregory Sin and Teresa Gonzalez Diaz and Ahmed Farahat and Chetan Gupta}},
      year={2025},
      eprint={2510.00259},
      note={arxiv:2510.00259}, 
}

@misc{claudesonnet,
  title        = {Claude Sonnet},
  author       = {{Anthropic}},
  year         = {2026},
  howpublished = {\url{https://www.anthropic.com/claude/sonnet}},
  note         = {Accessed: 2026-05-27}
}

@misc{openclaw,
  title        = {OpenClaw: Open-Source Agentic AI Framework},
  author       = {{OpenClaw Contributors}},
  year         = {2026},
  howpublished = {\url{https://github.com/openclaw/openclaw}},
  note         = {GitHub repository, Accessed: 2026-05-27}
}

\end{document}